\documentclass{article}

\usepackage[preprint,nonatbib]{tackling_climate_workshop_style}

\usepackage[utf8]{inputenc}
\usepackage[T1]{fontenc}
\usepackage[hidelinks]{hyperref}   
\usepackage{url}
\usepackage{booktabs}
\usepackage{amsfonts}
\usepackage{amsmath}
\usepackage{nicefrac}
\usepackage{microtype}
\usepackage{graphicx}
\usepackage{xcolor}
\usepackage{caption}

\graphicspath{{figures/}}
\title{Too Rare to Learn: Prescribed Cyclone Tracks\\Degrade a Bay of Bengal Ocean Emulator}

\author{%
  Sumaiya Islam \\
  Department of Software Engineering \\ University of Dhaka \\
  Dhaka, Bangladesh \\
  \texttt{bsse1446@iit.du.ac.bd} \\
}

\begin{document}

\maketitle

\begin{abstract}
Neural ocean emulators are being proposed for regional forecasting in cyclone-exposed
coastal seas, and a natural design choice is to hand the network the cyclone as a
prescribed input. We test that choice in the Bay of Bengal and find it harmful. We withhold
15 whole cyclones spanning 65 to 150 kt from GLORYS12 reanalysis and compare two U-Nets
that are identical except for four prescribed cyclone-track channels. Across three seeds the
ocean-only model beats persistence in every run and the storm-conditioned model loses to it
in every run, with the two skill ranges disjoint ($p = 3.1\times10^{-5}$, paired across
storms). The cause is exposure frequency rather than signal content: the channels are
non-zero on only 7.9\% of training days, so they are out of distribution the moment they
activate. The extra error falls inside the prescribed storm footprint, and replacing the real
cyclone map with a no-storm map at inference improves held-out storm forecasts by 7.5 to
16.4\% in every seed. The conditioned network has
learned a response to a rare signal that is confidently wrong.
\end{abstract}

\section{Introduction}

Coastal ocean forecasts matter most during tropical cyclones, and nowhere more than in
the Bay of Bengal, which produces a small share of the world's tropical cyclones but
most of the deadliest ones on record \cite{singh2022review}. Sea surface temperature sets the enthalpy available to an approaching storm and the cold
wake it leaves is a classical problem \cite{price1981upper}; in this basin sea surface
salinity controls the barrier layer that suppresses that cooling
\cite{singh2022review,thadathil2007barrier,vinayachandran2002barrier}. A next-day emulator of surface temperature
and salinity is therefore useful for coastal warning, provided it stays accurate on
cyclones it has never seen.

Recent neural ocean forecast systems predict the ocean from its own past state and
nothing else. GLONET, the operational neural system at Mercator Ocean, forecasts the
next day from the two preceding ocean states with no atmospheric forcing input
\cite{elaouni2024glonet}, and similar designs are used for global ocean emulation \cite{dheeshjith2025samudra},
eddy-resolving global forecasting \cite{wang2024xihe} and Bay of Bengal salinity
\cite{pasula2026bias}.
This leaves the network no way to know a cyclone is arriving, so its behaviour under an
unseen storm is fixed by what it inferred from ordinary days.

Prescribing the storm is the obvious remedy, and the literature brackets the question
without answering it. Cui et al. predict the cyclone-induced sea surface temperature response in the northwest
Pacific from track characteristics and pre-storm ocean properties, and report high skill
\cite{cui2023predicting}; their dataset is built entirely from cyclone cases, so a storm
appears in every sample. At the other extreme, AI weather models have transformed medium-range forecasting
\cite{lam2023graphcast,bi2023pangu} while under-predicting the peak amplitude of severe
storms \cite{charltonperez2024ciaran}, and Sun et al. show that FourCastNet
\cite{pathak2022fourcastnet} trained without Category 3 to 5 cyclones cannot forecast
Category 5 storms, without proposing a mechanism \cite{sun2025gray}. An operational regional emulator sits between these settings, with a cyclone channel
that is informative but rare. To our knowledge that intermediate regime has not been
measured.

This paper measures it, and finds that rare conditioning is worse than none. We withhold 15
whole Bay of Bengal cyclones with buffers and train two U-Nets differing only in four
prescribed cyclone-track channels, a gap of 864 parameters out of roughly 101{,}000. The
conditioned model loses to persistence on held-out cyclones in all six runs; the
unconditioned model beats it in all six. We trace the failure to the sparsity of the conditioning signal rather than its content:
the channels are non-zero on 7.9\% of training days, and the penalty concentrates on the
moments and the places where they activate. Our
contributions are an event-disjoint Bay of Bengal protocol, with 15 held-out cyclones spanning 65 to 150
kt and both monsoon seasons and checkpoint selection on storm-free windows so it cannot
leak storm-regime information into the test claim (Appendix~\ref{app:storms}); a
replicated negative result whose skill ranges do not overlap across three seeds or
either checkpoint policy; and a mechanism, supported by four diagnostics including an
inference-time intervention that recovers the lost skill, identifying channel occupancy
rather than signal content as the cause. 
\section{Experimental setup}

\begin{figure}[!htbp]
\centering
\includegraphics[width=0.76\linewidth]{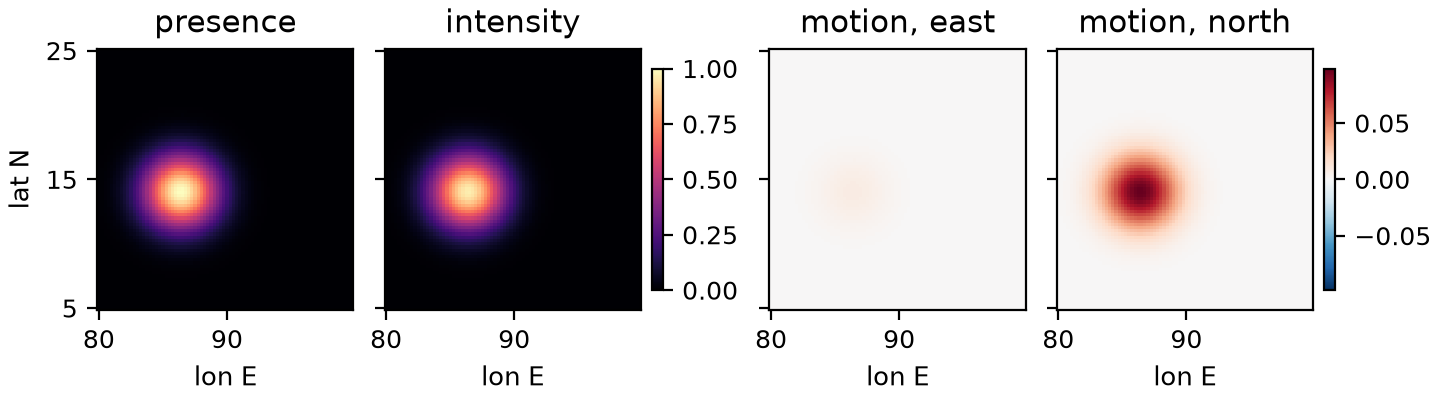}
\caption{The four prescribed cyclone channels for one day of Amphan, 18 May 2020: smooth
track proxies carrying no ocean state. Amphan moved almost due north, hence the faint
eastward panel.}
\label{fig:channels}
\end{figure}

\paragraph{Data and cyclone channels.} We use GLORYS12V1 daily-mean reanalysis \cite{lellouche2021glorys} regridded to
$0.25^{\circ}$ over $5$--$25^{\circ}$N, $80$--$99.75^{\circ}$E, an $81 \times 80$ grid
spanning 12{,}227 days from 1993 to June 2026. The state is surface temperature and
salinity, standardised per variable. Each IBTrACS \cite{knapp2010ibtracs} best-track fix
is rendered as a Gaussian kernel with $\sigma = 2^{\circ}$, giving the four daily fields
in Figure~\ref{fig:channels}: presence, intensity and two motion components
(Appendix~\ref{app:config}). The fields are taken at the forecast target day, so the conditioned model is handed the
storm it is forecasting into: the most favourable case for conditioning.

\paragraph{Event-disjoint splits.} A random or chronological split places days from the same cyclone in both training and
test data, so we withhold whole events instead. Fifteen severe cyclones are held out by round-robin across intensity terciles crossed with monsoon season, so the test set spans 65 to
150 kt in both seasons rather than only famous storms (Appendix~\ref{app:storms}). Each
is buffered by 7 days, giving 272 evaluation windows. Checkpoint selection uses 33
storm-free 21-day blocks, so it cannot tune the reported weights to the regime under
test.

\paragraph{Models, training and metric.} Both arms use the same U-Net \cite{ronneberger2015unet}, width 24, predicting the next-day state from three days of history. The ocean-only arm takes six input channels and
the storm-conditioned arm ten; at 100{,}706 and 101{,}570 parameters they differ in
capacity by 0.9\%, so the comparison isolates the conditioning signal. Training uses masked MSE over ocean points and AdamW \cite{loshchilov2019adamw}, for
seeds 1, 2 and 3; Appendix~\ref{app:config} gives the full configuration. We report both the quiet-selected checkpoint (\emph{best}) and the final epoch
(\emph{final}). Skill is measured against persistence, $S = 1 - \mathrm{RMSE}_{\text{model}} /
\mathrm{RMSE}_{\text{pers}}$, which repeats the most recent observed state. Persistence is the strongest classical baseline at this lead
\cite{rasp2024weatherbench}: climatology scores $-2.84$ and damped persistence only
$+0.010$ (Appendix~\ref{app:baselines}). Pooled skill uses the standardised state over both
variables, and per-variable RMSE is given in $^{\circ}$C and PSU.

\section{Results}

\paragraph{Conditioning flips the sign of the skill.} Table~\ref{tab:pooled} gives pooled
skill on the 272 held-out cyclone windows. Every ocean-only run beats persistence and every
storm-conditioned run loses to it, under both checkpoint policies. The ranges $[+0.048, +0.161]$ and $[-0.477, -0.075]$ do not overlap. Paired across the
15 storms, the ocean-only arm exceeds the conditioned arm at $p = 3.1\times10^{-5}$ by a
Wilcoxon signed-rank test, and separately in every seed at $p = 0.011$,
$9.2\times10^{-5}$ and $3.1\times10^{-5}$, so the ordering does not rest on the seed 3
outlier (Appendix~\ref{app:stats}).

\begin{table}[!htbp]
\caption{Skill against persistence on 272 windows from 15 held-out cyclones, higher is
better.}
\label{tab:pooled}
\centering
\footnotesize
\begin{tabular}{llccc c}
\toprule
Arm & Checkpoint & Seed 1 & Seed 2 & Seed 3 & Storms beaten, by seed (of 15) \\
\midrule
Ocean-only & best  & $+0.125$ & $+0.153$ & $+0.161$ & 12 / 13 / 14 \\
Ocean-only & final & $+0.048$ & $+0.153$ & $+0.161$ & 11 / 13 / 14 \\
Storm-conditioned & best  & $-0.206$ & $-0.194$ & $-0.293$ & 6 / 6 / 4 \\
Storm-conditioned & final & $-0.075$ & $-0.078$ & $-0.477$ & 6 / 6 / 0 \\
\bottomrule
\end{tabular}
\end{table}

\begin{figure}[!htbp]
\centering
\includegraphics[width=0.80\linewidth]{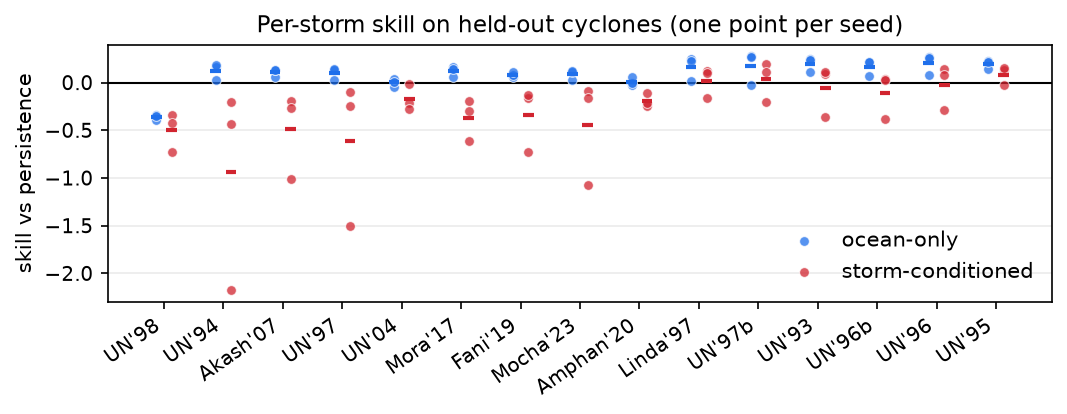}
\caption{Per-cyclone skill against persistence, one point per seed, bar at the seed
mean, ordered by persistence difficulty.}
\label{fig:perstorm}
\end{figure}

\paragraph{The result holds storm by storm, with an honest exception.} Ocean-only is
above persistence on 11 to 14 of the 15 cyclones depending on seed, losing on an unnamed
1998 storm in all three (Figure~\ref{fig:perstorm}). Conditioning is not uniformly catastrophic: seeds 1 and 2 reach persistence on 6
cyclones, none among the four most intense.

\paragraph{The recoverable skill is entirely in salinity.} Separating the state variables in
physical units, averaged over seeds at the final epoch, persistence gives 0.141$^{\circ}$C and
0.226 PSU, ocean-only 0.168$^{\circ}$C and 0.190 PSU, and storm-conditioned 0.318$^{\circ}$C and
0.236 PSU. Both arms are worse than persistence at temperature, so the ocean-only advantage is
salinity alone, improving on persistence by 0.036 PSU where the conditioned arm does not
improve at all. Salinity is the Bay-of-Bengal-specific variable, set by Ganges and
Brahmaputra freshwater and controlling the barrier layer, and it dominates the pooled
metric because its normalised error is the larger of the two
(Appendix~\ref{app:variables}).

\section{Why conditioning hurts}

\paragraph{The conditioning signal is rare.} The four cyclone channels carry a non-zero value on 885 of the 11{,}195 training days,
or 7.9\%; on the other 92.1\% the network sees a constant no-storm field
(Appendix~\ref{app:occupancy}). Any response it learns to a storm map is estimated from a small, unrepresentative slice
of training.

\begin{figure}[!htbp]
\centering
\includegraphics[width=0.80\linewidth]{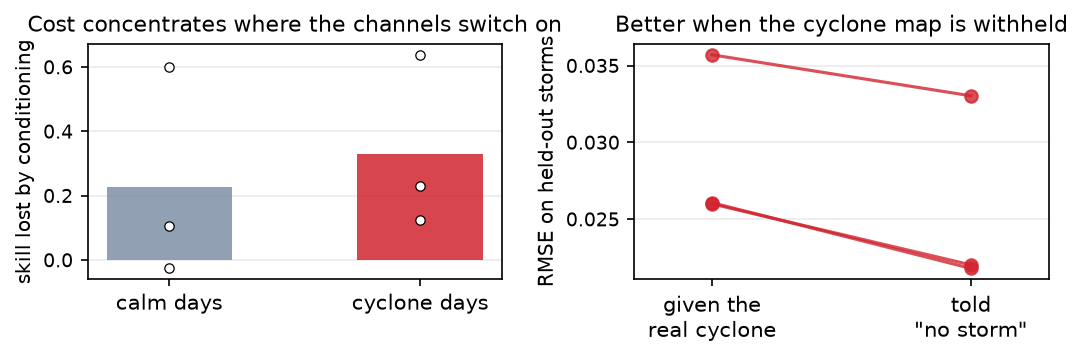}
\caption{Left: skill lost to conditioning, by regime, one point per seed. Right: RMSE on
held-out cyclone days given the real cyclone map and given a no-storm map. The channels
are active on 7.9\% of training days.}
\label{fig:mech}
\end{figure}

\paragraph{The penalty concentrates where the channels switch on.} Evaluating both arms on held-out cyclone days and storm-free days separately, the cost
of conditioning is larger on cyclone days in every seed, most cleanly for seed 1 at
$-0.026$ on calm days against $+0.123$ on storm days (Figure~\ref{fig:mech} left,
Appendix~\ref{app:regime}). Conditioning is mildly beneficial while the channels are off
and clearly harmful once they switch on.

\paragraph{Silencing the cyclone map improves storm forecasts.} We overwrite the four cyclone channels at inference with the value encoding ``no storm''
(Appendix~\ref{app:bugs}). On held-out cyclone days this improves RMSE in every seed, by
15.3\%, 16.4\% and 7.5\% (Figure~\ref{fig:mech}, right): the model forecasts a cyclone
better when told the cyclone is not there. An uninformative channel would leave the forecast unchanged; one triggering a wrong
learned response makes it worse.

\paragraph{The penalty falls where the channels are.} Splitting held-out cyclone days into the prescribed storm footprint and everything else,
conditioning costs $+70.6\%$, $+97.8\%$ and $+101.5\%$ of salinity RMSE inside against $+6.2\%$,
$+15.8\%$ and $+48.5\%$ outside, a ratio of 11.3, 6.2 and 2.1 (Appendix~\ref{app:spatial}). This answers
the natural objection that the conditioned arm is simply under-trained or too small at
101k parameters: either would spread the extra error uniformly, and neither would gain
accuracy when the cyclone map is withheld.  Together the four diagnostics point to
exposure frequency rather than signal content: a
channel active on 7.9\% of training days is out of distribution precisely when it
activates. Autoregressive emulators reach that state only after long rollouts
\cite{pedersen2025thermalizer}, whereas here the input starts there, and Cui et al.
\cite{cui2023predicting} see no such failure because their conditioning is never sparse.

\section{Conclusions and future work}

A prescribed cyclone track made a regional ocean emulator worse than persistence on
cyclones it had never seen, while the same network without it beat persistence. The
failure tracks the rarity of the conditioning signal rather than its content, so a
conditioning channel should be judged on how often it is exercised during training, not
only on how informative it is when present. The claim is scoped to track-derived conditioning at a one-day lead
(Appendices~\ref{app:limits} and~\ref{app:seed3}). Future work will add a third arm driven by dense ERA5 forcing \cite{hersbach2020era5}
and extend to multi-day rollouts.

\section*{Climate impact statement}

Coastal Bangladesh and eastern India depend on short-range upper-ocean forecasts during
cyclone season. We show that an intuitive way to make a regional emulator storm-aware
instead pushes it below a trivial baseline on the events that matter most, where calm-day validation would miss the problem. The protocol and diagnostic transfer to any
emulator with a rare conditioning input.

\bibliographystyle{plain}

\appendix

\section{Held-out cyclones}
\label{app:storms}

The 15 test cyclones (Table~\ref{tab:storms}) were chosen deterministically by round-robin
across intensity terciles crossed with monsoon season, not by fame, so that the test set can
answer \emph{when} conditioning fails rather than only \emph{whether} it fails. IBTrACS leaves
many Bay of Bengal storms unnamed; labels are derived per storm identifier with a month suffix
where a name and year still collide, and the build asserts uniqueness so that two distinct
cyclones can never be merged into one reported row.

\begin{table}[h]
\caption{The 15 held-out cyclones, ordered by intensity. Pre-monsoon is March to June,
post-monsoon September to December.}
\label{tab:storms}
\centering\footnotesize
\begin{tabular}{llccr}
\toprule
Name & Year & Month & Season & Max wind (kt) \\
\midrule
Fani & 2019 & Apr & pre & 150 \\
Amphan & 2020 & May & pre & 145 \\
Mocha & 2023 & May & pre & 145 \\
Unnamed & 1994 & Apr & pre & 125 \\
Unnamed & 1996 & Nov & post & 115 \\
Unnamed & 1997 & May & pre & 115 \\
Unnamed & 1995 & Nov & post & 105 \\
Mora & 2017 & May & pre & 80 \\
Unnamed & 1993 & Nov & post & 75 \\
Unnamed & 1996 & Nov & post & 75 \\
Unnamed & 1998 & May & pre & 70 \\
Linda & 1997 & Nov & post & 65 \\
Unnamed & 1997 & Sep & post & 65 \\
Unnamed & 2004 & May & pre & 65 \\
Akash & 2007 & May & pre & 65 \\
\bottomrule
\end{tabular}
\end{table}

\begin{figure}[h]
\centering
\includegraphics[width=0.92\linewidth]{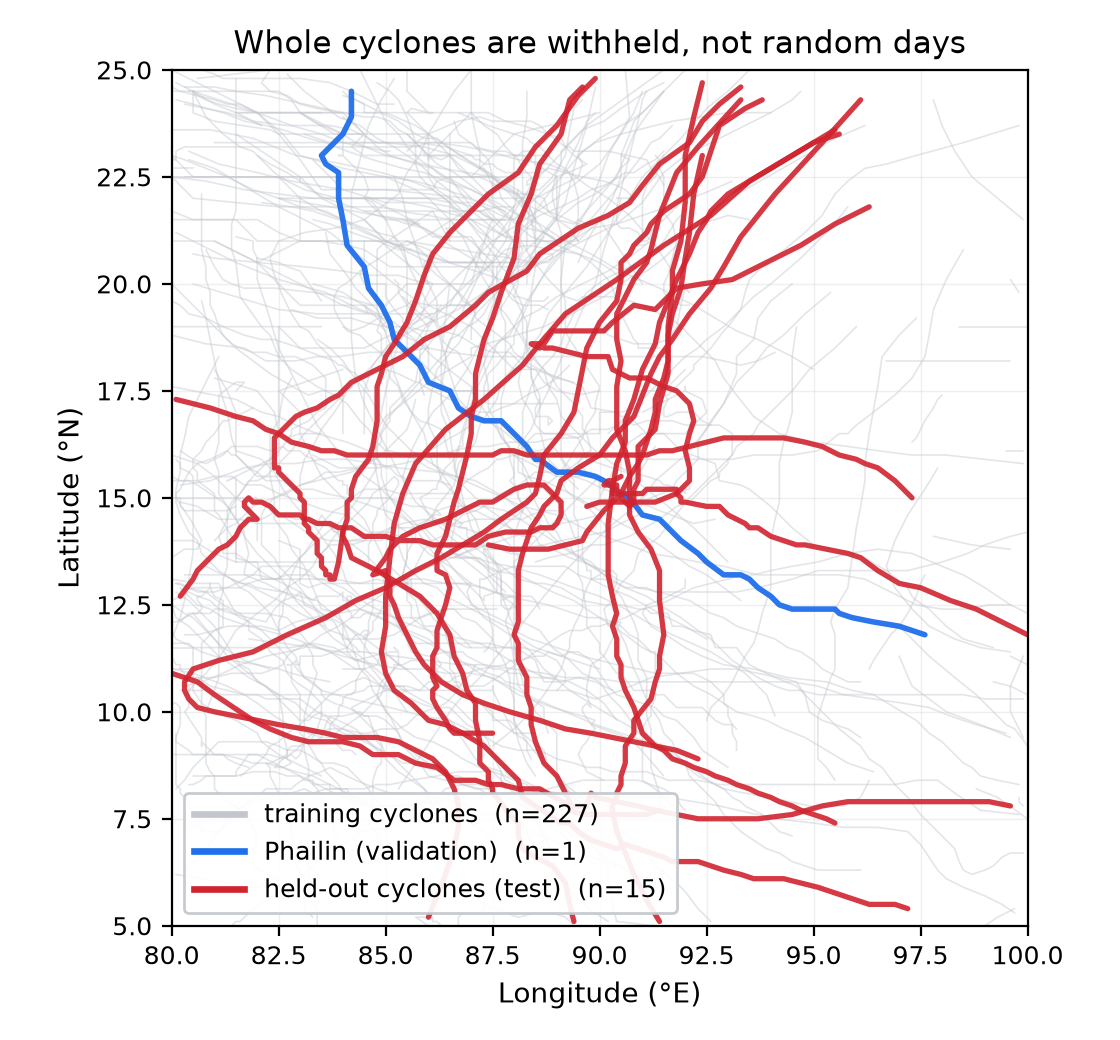}
\caption{The event-disjoint split, drawn from IBTrACS tracks: 227 training cyclones,
Phailin held for validation, and the 15 test cyclones. Whole events are withheld with
7-day buffers, so no day from a test storm, or from the week either side of it, appears
in training.}
\label{fig:split}
\end{figure}

\section{Per-variable results}
\label{app:variables}

Table~\ref{tab:vars} separates temperature and salinity by seed in physical units. Both arms
are worse than persistence at temperature at a one-day lead, in every run. The ocean-only
advantage is salinity alone, and it is present in all three seeds.

\begin{table}[h]
\caption{Pooled RMSE on held-out cyclones in physical units, final-epoch checkpoints, per seed.
Lower is better; bold beats persistence.}
\label{tab:vars}
\centering\footnotesize
\begin{tabular}{lcccccc}
\toprule
& \multicolumn{3}{c}{SST RMSE ($^{\circ}$C) $\downarrow$} & \multicolumn{3}{c}{SSS RMSE (PSU) $\downarrow$} \\
\cmidrule(lr){2-4}\cmidrule(lr){5-7}
Model & Seed 1 & Seed 2 & Seed 3 & Seed 1 & Seed 2 & Seed 3 \\
\midrule
Persistence & \multicolumn{3}{c}{0.141} & \multicolumn{3}{c}{0.226} \\
Ocean-only & 0.174 & 0.163 & 0.166 & \textbf{0.208} & \textbf{0.183} & \textbf{0.180} \\
Storm-conditioned & 0.248 & 0.244 & 0.463 & 0.222 & 0.225 & 0.261 \\
\bottomrule
\end{tabular}
\end{table}

\begin{figure}[h]
\centering
\includegraphics[width=0.88\linewidth]{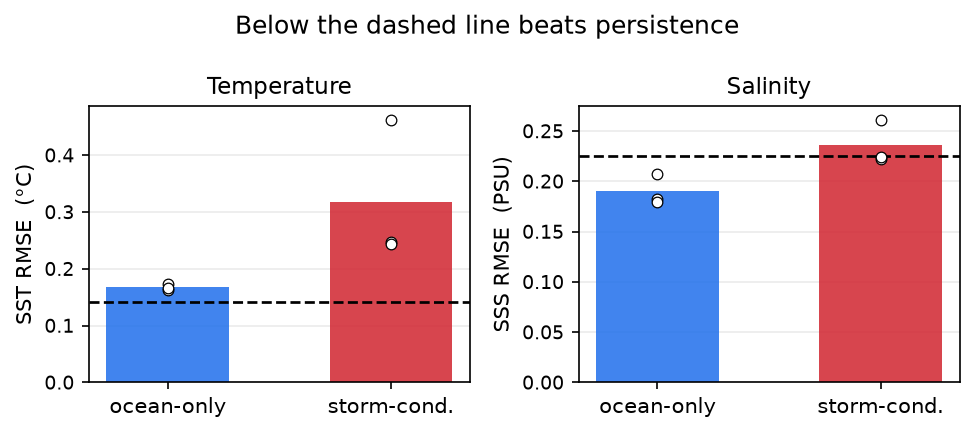}
\caption{Where the skill comes from. Bars are seed means, points individual seeds. Only
ocean-only salinity falls below the persistence line.}
\label{fig:vars}
\end{figure}

\section{Cost of conditioning by regime}
\label{app:regime}

Table~\ref{tab:regime} gives the numbers behind the left panel of Figure~\ref{fig:mech}. The
skill penalty is computed against the persistence baseline for that regime, so calm and cyclone
days are each measured against their own reference. Seed 1 is the cleanest demonstration:
conditioning is mildly \emph{beneficial} on calm days and clearly harmful on storm days, which
is what an input that goes out of distribution on activation predicts, and what a merely worse
model does not.

\begin{table}[h]
\caption{Ocean-only skill minus storm-conditioned skill, by regime. Positive means conditioning
hurts.}
\label{tab:regime}
\centering\footnotesize
\begin{tabular}{lccc}
\toprule
Regime & Seed 1 & Seed 2 & Seed 3 \\
\midrule
Calm days & $-0.026$ & $+0.105$ & $+0.599$ \\
Cyclone days & $+0.123$ & $+0.232$ & $+0.638$ \\
\bottomrule
\end{tabular}
\end{table}

\section{Channel occupancy}
\label{app:occupancy}

\begin{figure}[h]
\centering
\includegraphics[width=\linewidth]{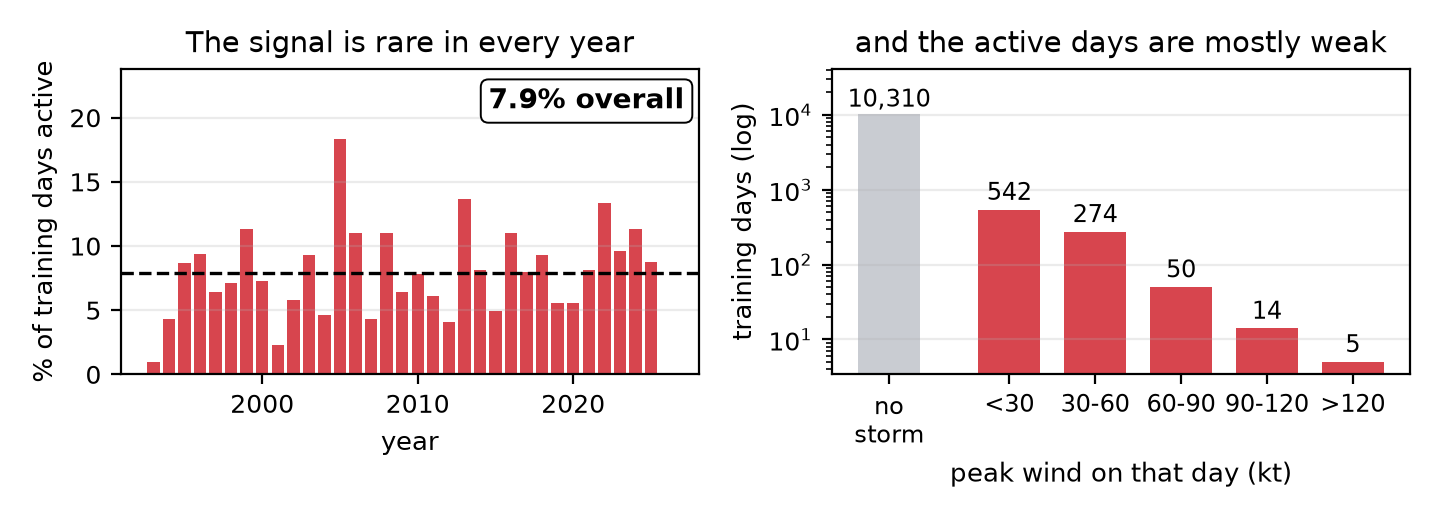}
\caption{Occupancy of the cyclone channels over the 11{,}195 training days. Left: the fraction
of days carrying any signal, by year, against the 7.9\% overall rate. Right: training days by
the peak wind on that day, on a log scale. The network sees no storm on 10{,}310 days and a
storm above 120 kt on five.}
\label{fig:occupancy}
\end{figure}

The mechanism turns on how often the network is exposed to the conditioning signal, so
Figure~\ref{fig:occupancy} counts that directly over the training split, with all held-out
days excluded. The cyclone channels are non-zero on 885 of 11{,}195 training days, 7.9\%, and
that rate is stable across the record rather than concentrated in a few years.

The right panel sharpens the problem. Exposure is not only rare, it is heavily skewed toward
weak storms: of the 885 active training days, 542 carry a peak wind below 30 kt and only 19
carry one above 90 kt. \textbf{Just 5 training days, 0.045\% of the split, contain a storm above 120 kt.} The test set contains four cyclones at that intensity, Fani, Amphan, Mocha and the
unnamed 1994 storm, and seven above 90 kt. The conditioned model is therefore asked to apply a
response function it estimated from a handful of days, which is the same exposure-frequency
argument Sun et al. \cite{sun2025gray} make for unseen intense cyclones, arriving here through
the conditioning input rather than the target state.

\section{Channel construction and training configuration}
\label{app:config}

Each best-track fix becomes a Gaussian kernel with $\sigma = 2^{\circ}$; intensity is
that kernel scaled by maximum wind normalised by 150 kt, and the two motion channels by
translation velocity normalised by 30\,m\,s$^{-1}$ and clipped to $[-1, 1]$. Both arms
use an identical U-Net: one downsampling level, width 24, $3\times3$ convolutions with
GroupNorm and GELU, bilinear upsampling, and a $1\times1$ head onto two output channels. Input is
three days of history, six channels for the ocean-only arm and ten for the storm-conditioned arm.
Parameter counts are 100{,}706 and 101{,}570, a difference of 864 or 0.9\%, so capacity is not a
confound. Training uses masked MSE over ocean points only, AdamW at learning rate $10^{-3}$ with
weight decay $10^{-4}$, batch size 32, 30 epochs, on a single GPU. Each run takes about 25
minutes. Inputs are standardised per variable using training-split statistics stored in the
checkpoint, so evaluation cannot silently use different normalisation from training.

Checkpoint selection uses 33 storm-free 21-day blocks (594 windows), each at least 7 days from
any cyclone. Selecting on storm windows instead would tune the reported weights to the very
regime under test. Cyclone Phailin (2013) is monitored during training but never promoted to the
test set.

\section{Limitations}
\label{app:limits}

The cyclone channels carry position, intensity and motion but no ocean state and no wind stress
or heat flux, so this experiment does not show that dense physical forcing would fail the same
way, and the sparsity argument does not transfer to a field that is present every day.
Evaluation is at a one-day lead; the 3- and 7-day rollouts the protocol specifies are not run.
Both arms are worse than persistence at temperature, so the positive result is confined to
salinity. The study covers one basin, one architecture and one reanalysis; repeating it on ORAS5
\cite{zuo2019oras5} would test whether the effect is a property of GLORYS12. The cyclone maps are
built from best-track data, giving the conditioned model perfect knowledge of the storm it is
forecasting into, which is more favourable than an operational setting would be.

\section{Seed 3 of the conditioned arm}
\label{app:seed3}

Seed 3 of the storm-conditioned arm is a poorly converged outlier: $-0.477$ pooled against about
$-0.076$ for seeds 1 and 2, and below persistence on all 15 storms rather than 9. It is worse
\emph{everywhere}, including on calm days, where seeds 1 and 2 are close to the ocean-only arm.
That pattern is a failed optimisation rather than a stronger version of the conditioning effect.

It is also the run that was killed and restarted after the machine went into swap, so it is not
the trajectory that would otherwise have completed; CUDA non-determinism means a restart with
the same seed does not reproduce the same run. We report it rather than dropping it, and we
report seeds individually throughout, because a mean across the three describes none of them.
The direction of the result does not depend on it: seeds 1 and 2 alone give a storm-conditioned
range of $[-0.078, -0.075]$, still disjoint from the ocean-only range.

\section{Where the penalty falls}
\label{app:spatial}

A model that is simply worse should be worse everywhere. A model that has learned a wrong
response to the cyclone map should be worse where that map is non-zero. Table~\ref{tab:spatial}
splits every held-out cyclone day into the prescribed storm footprint, taken as presence above
0.2, and everything else, and scores both arms on each region separately. Of the 272 test
windows, 112 carry a footprint above that threshold; the remainder fall in the 7-day buffers
either side of a cyclone and have no footprint to split on.

The penalty inside the footprint is 11.3, 6.2 and 2.1 times the penalty outside it for
seeds 1 to 3, so it is larger inside in every seed, though by a margin that narrows as
the run degrades.
This is the clearest evidence that the conditioned arm has not merely failed to train: an arm
that were under-trained or under-parameterised would carry its extra error uniformly, and would
not gain accuracy when the cyclone map is replaced by a no-storm map at inference.

\begin{table}[h]
\caption{Salinity RMSE (PSU) on the 112 held-out cyclone days, inside and outside the
prescribed storm footprint, final-epoch checkpoints. Penalty is the storm-conditioned arm
relative to the ocean-only arm in the same region.}
\label{tab:spatial}
\centering\footnotesize
\begin{tabular}{lcccccc}
\toprule
& \multicolumn{3}{c}{Inside footprint} & \multicolumn{3}{c}{Outside footprint} \\
\cmidrule(lr){2-4}\cmidrule(lr){5-7}
& Seed 1 & Seed 2 & Seed 3 & Seed 1 & Seed 2 & Seed 3 \\
\midrule
Ocean-only & 0.291 & 0.276 & 0.272 & 0.216 & 0.191 & 0.187 \\
Storm-conditioned & 0.497 & 0.545 & 0.548 & 0.230 & 0.222 & 0.277 \\
\midrule
Penalty & $+70.6\%$ & $+97.8\%$ & $+101.5\%$ & $+6.2\%$ & $+15.8\%$ & $+48.5\%$ \\
\bottomrule
\end{tabular}
\end{table}

\begin{figure}[h]
\centering
\includegraphics[width=\linewidth]{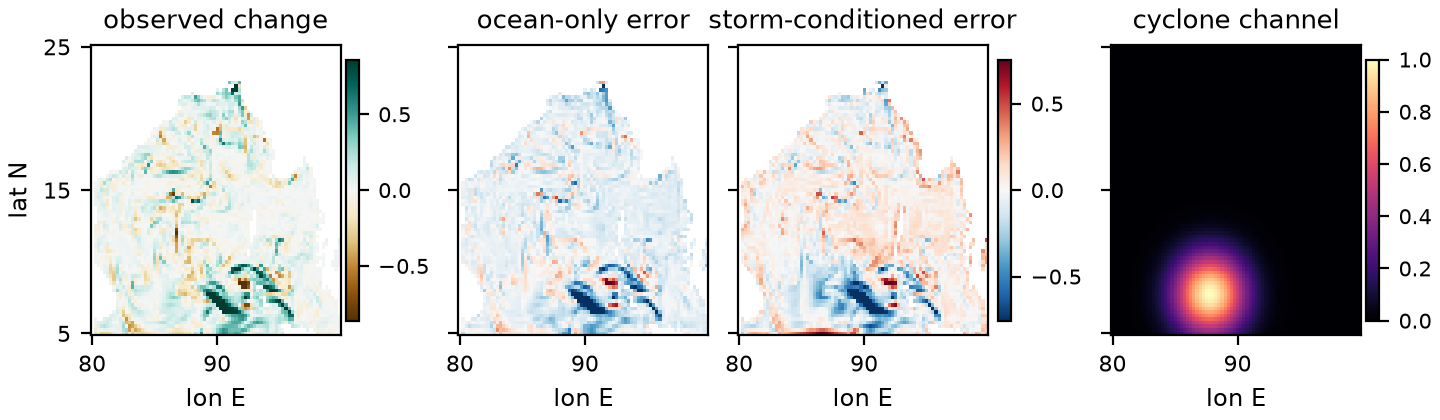}
\caption{One day of Cyclone Fani, 28 April 2019, seed 2. Left to right: the observed one-day
salinity change (PSU), the ocean-only error, the storm-conditioned error on the same colour
scale, and the cyclone channel the conditioned arm was given. The conditioned arm carries a
broad error across the southern basin, where the channel is active. Fani is the held-out
cyclone whose skill gap is closest to the median among those showing the reported sign pattern,
and the day drawn is the one with the strongest prescribed signal; neither was chosen by
inspecting the errors.}
\label{fig:forecast}
\end{figure}

\section{Classical baselines}
\label{app:baselines}

Persistence is the only baseline in the main text, so we check that it is not a
straw man. Table~\ref{tab:baselines} adds day-of-year climatology, computed over training days
only and smoothed with a 15-day window, and damped persistence,
$\mathrm{clim} + \alpha(\mathrm{yesterday} - \mathrm{clim})$, with $\alpha$ fitted on the 594
storm-free validation windows and never on the test set. Scores use the same aggregation as
Table~\ref{tab:pooled}.

At a one-day lead the fitted $\alpha$ is 0.97, meaning the best classical blend is almost pure
persistence, and climatology alone is nearly four times worse. Damped persistence improves on
persistence by 0.010 in skill. The ocean-only arm's $+0.048$ to $+0.161$ therefore clears the
strongest classical baseline as well as the simplest one, and the storm-conditioned arm's
$-0.075$ to $-0.477$ is below all three.

\begin{table}[h]
\caption{Classical baselines on the same 272 held-out cyclone windows, standardised state,
lower RMSE is better.}
\label{tab:baselines}
\centering\footnotesize
\begin{tabular}{lcc}
\toprule
Baseline & RMSE $\downarrow$ & Skill vs persistence \\
\midrule
Climatology & 0.0929 & $-2.842$ \\
Persistence & 0.0242 & $0.000$ \\
Damped persistence ($\alpha = 0.97$) & \textbf{0.0239} & $+0.010$ \\
\bottomrule
\end{tabular}
\end{table}

\section{Statistical tests}
\label{app:stats}

Skill is paired by storm, so we use a Wilcoxon signed-rank test over the 15 held-out cyclones
rather than assuming normality across so few events. Pooling seeds, the ocean-only arm exceeds
the storm-conditioned arm at $p = 3.1\times10^{-5}$ (one-sided, $n = 15$). Run separately the
test gives $p = 0.011$, $9.2\times10^{-5}$ and $3.1\times10^{-5}$ for seeds 1, 2 and 3, so the
ordering holds in each seed on its own and does not depend on the seed 3 outlier. Against zero,
ocean-only skill is positive at $p = 0.0042$ and storm-conditioned skill negative at
$p = 1.0\times10^{-3}$.

As a further robustness check, a preliminary version of this experiment held out only
three cyclones and gave $+0.051$ pooled skill for the ocean-only arm against $+0.161$
here, so the effect grows rather than shrinks with a larger held-out set.

\section{Correctness checks}
\label{app:bugs}

The comparison rests on a persistence baseline and on two models being fed the inputs they were
trained on, so we list the errors found and fixed while building the pipeline. Each would have
produced a confident but wrong number.

\begin{itemize}
\setlength{\itemsep}{1pt}
\item The persistence baseline indexed the most recent state with the wrong stride and read past
the end of the input tensor. The correct stride is the number of state variables, not the number
of input channels.
\item The test dataset was always built without the cyclone channels, so a ten-channel checkpoint
met six-channel data. The dataset is now built to match the checkpoint, with an explicit channel
assertion.
\item Silencing the cyclone channels by writing $0.0$ into standardised space encodes a
mean-intensity storm everywhere, not the absence of a storm. The correct silent value is
$(0-\mu)/\sigma$. Before the fix, the silencing experiment read $+57\%$ and pointed at the
opposite conclusion. As a control, the same substitution on calm days changes RMSE by 0.0\%,
since those days already carry the no-storm value.
\item Channel occupancy measured on standardised values reports 100\% active, because ``no
storm'' is not zero after standardisation. On raw values it is 7.9\%. Counting over the
whole record rather than the training split alone also inflates it, to 8.2\%.
\item Per-storm results keyed on storm name would have merged seven distinct unnamed cyclones
into shared rows; 1996 and 1997 each contain two unnamed severe storms.
\item Evaluation keyed results on the checkpoint filename, which is identical across seeds, so
each seed silently overwrote the previous one and only 4 of 12 checkpoints were reported.
\end{itemize}

\section{Data and code availability}
\label{app:availability}

Both inputs are public and require no request beyond a free account.

\textbf{Ocean state.} GLORYS12V1 daily-mean reanalysis \cite{lellouche2021glorys} from the
Copernicus Marine Service, product
\texttt{GLOBAL\_\allowbreak MULTIYEAR\_\allowbreak PHY\_\allowbreak 001\_\allowbreak 030},
dataset
\texttt{cmems\_\allowbreak mod\_\allowbreak glo\_\allowbreak phy\_\allowbreak my\_\allowbreak 0.083deg\_\allowbreak P1D-m\_\allowbreak 202311}.
We take the shallowest level
(0.494\,m) of \texttt{thetao} and \texttt{so} over $5$--$25^{\circ}$N,
$80$--$100^{\circ}$E for 1993-01-01 to 2026-06-23, then regrid to $0.25^{\circ}$.

\textbf{Cyclone tracks.} IBTrACS v04r01 \cite{knapp2010ibtracs}, North Indian Ocean subset
(\texttt{ibtracs.\allowbreak NI.\allowbreak list.\allowbreak v04r01.\allowbreak csv}), from
NOAA NCEI.

\textbf{Code.} The training, evaluation and diagnostic code will be released publicly on
acceptance. The grid, date range, event-disjoint split rule, architecture, parameter counts
and training hyperparameters needed to reproduce the experiment are given in
Appendix~\ref{app:config}, and the held-out cyclones are listed in
Appendix~\ref{app:storms}.

\end{document}